%% file: main.tex
\documentclass[journal,onecolumn]{IEEEtran}

\usepackage{cite}
\usepackage[pdftex]{graphicx}
\graphicspath{{figures/}}
\DeclareGraphicsExtensions{.pdf,.jpeg,.png}
\usepackage{algorithmic}
\usepackage{array}
\usepackage{url}
\usepackage{blindtext}
\usepackage{color}
\usepackage{todonotes}
\usepackage{caption}
\usepackage{subcaption}

\begin{document}

\title{Spatial Computing and Intuitive Interaction: Bringing Mixed Reality and Robotics Together}

\author{Jeffrey Delmerico, Roi Poranne, Federica Bogo, Helen Oleynikova, Eric Vollenweider, Stelian Coros, Juan Nieto, Marc Pollefeys
\thanks{J. Delmerico, F. Bogo, H. Oleynikova, E. Vollenweider, J. Nieto, and M. Pollefeys are with the Microsoft Mixed Reality and AI Lab, Z\"{u}rich, Switzerland. e-mail: \{firstname.lastname@microsoft.com\}}%
\thanks{R. Poranne and S. Coros are with the Computational Robotics Lab at ETH Z\"{u}rich, Switzerland. e-mail: \{roi.poranne, scoros\}@inf.ethz.ch}%
\thanks{R. Poranne is also with department of computer science at the University of Haifa, Israel. e-mail: roiporanne@cs.haifa.ac.il}%
\thanks{H. Oleynikova is now with NVIDIA. e-mail: holeynikova@nvidia.com }%
}

\markboth{IEEE Robotics and Automation Magazine}%
{Special Issue on Extended Reality in Robotics}
% after the title page when using the twoside option.https://www.overleaf.com/project/60c0bb3c8d5f7d316cce7575
% 

\maketitle

\IEEEpeerreviewmaketitle

\input{abstract}

\input{introduction}
\input{sharing_spatial_information}
\input{colocalization}
\input{embodiment}
\input{conclusion}
\input{acknowledgment}

\ifCLASSOPTIONcaptionsoff
  \newpage
\fi



\bibliographystyle{IEEEtran}
\bibliography{bib/IEEEabrv.bib,bib/references.bib}


% \end{IEEEbiography}


% % insert where needed to balance the two columns on the last page with
% %\newpage



%\vfill


\end{document}

%% file: abstract.tex
\begin{abstract}
Spatial computing---the ability of devices to be aware of their surroundings and to represent this digitally---offers novel capabilities in human-robot interaction.  
In particular, the combination of spatial computing and egocentric sensing on mixed reality devices enables them to capture and understand human actions and translate these to actions with spatial meaning, which offers exciting new possibilities for collaboration between humans and robots.
This paper presents several human-robot systems that utilize these capabilities to enable novel robot use cases: mission planning for inspection, gesture-based control, and immersive teleoperation. 
These works demonstrate the power of mixed reality as a tool for human-robot interaction, and the potential of spatial computing and mixed reality to drive the future of human-robot interaction.
\end{abstract}

%% file: introduction.tex
\section{Introduction}
\IEEEPARstart{S}{ince} the term was first coined almost twenty years ago~\cite{greenwold2003spatial}, the rise of virtual, augmented, and mixed reality technologies has highlighted the importance and applications of \textit{spatial computing} as a novel paradigm for interacting with spatial information through our devices.
Spatial computing refers to the digitization and modeling of the device's environment and the objects within it, such that the device has spatial context.
What distinguishes mixed reality from virtual realty is the ability to observe \textit{both} the physical and digital world simultaneously, with digital content aligned to the real spatial environment, and the ability to interact with both physical and digital objects.  
While augmented reality devices offer some of these capabilities through multi-touch screens, mixed reality is characterized by a more immersive visualization and interaction experience through a head-mounted display.

In the world of mobile robotics, spatial computing is a requirement for most operations.  
In order to navigate in an environment, avoid obstacles, and perform useful functions, mobile robots need to build and use a digital representation of their understanding of the environment.
Often this takes the form of a map, where the robot maintains an estimate of where it is in space, as well as the structure of the environment.
But more broadly, this representation can be a \textit{digital twin} of the environment, where any information about the world that has spatial meaning can be embedded in a digital framework intended to capture information about the space, and the digital devices in it, as accurately and completely as possible.

The trend in devices taking advantage of spatial computing is often called the \textit{Third Wave of Personal Computing}, after the first two waves where desktops provided access to home computing, and then mobile devices initially made this computing power ubiquitous and portable, but without any awareness of the space they are in.
Mobile robots and mixed reality (MR) devices thus have significant synergies, since they require many of the same spatial computing capabilities, regardless of whether they are designed to be used by a human, or to behave autonomously. 

This alignment of spatial understanding between human-oriented MR devices and robots provides an opportunity to unlock new modes of interaction between humans and machines.  
In particular, by sharing these spatial representations or digital twins between human and robotic devices, we can enable all of the agents involved to leverage that information for greater capabilities.
This enables them to \textit{do more}, and provides a common understanding so that humans and robots can all work together better and more efficiently through natural and intuitive interaction and collaboration.

Additionally, by leveraging the egocentric sensing and immersive visualization of MR devices, we can also provide embodied teleoperation experiences for remote devices. 
In this scenario, the user projects their actions to a remote robot while using the robot's spatial understanding to provide immersive feedback on the task at hand.

In this paper, we aim to show our efforts in this domain, with three works that illustrate how mixed reality can enable more flexible collaboration and even new types of interactions between humans and robots.
Section~\ref{sec:mission_planning} details a system in which a user can plan autonomous inspection missions for a robot, which is made possible by sharing a persistent spatial reference frame in which these missions are defined. 
We then describe an approach for using the egocentric sensing and human understanding capabilities of a mixed reality headset to control a robot using gestures in Section~\ref{sec:colocalization}, which builds on the previous section with simultaneous colocalization of the robot and human.
Finally, Section~\ref{sec:immersive_teleop} removes the colocalization aspect and instead explores the use of immersive mixed reality experiences to provide the user with a remote robot's spatial understanding, while retargeting the MR device's egocentric sensing to remotely teleoperate the robot. 

%% file: sharing_spatial_information.tex
\section{Sharing Spatial Information}
\label{sec:mission_planning}
We first consider a scenario in which devices share spatial information in a temporally decoupled fashion.
Here, we demonstrate that spatial data can be persisted over time by defining it with respect to world-locked reference coordinate systems.  
The general workflow begins with one device defining a reference coordinate system by building a visual map of the environment.
Spatial information is then anchored to this reference frame, which is fixed with respect to the space, enabling it to persist in that location over time.
Another device, or the first device at a later time, can then relocalize to this map, recover the reference coordinate system, and access the stored information in the same place in the world where it was defined.

In this section, we show how this type of workflow can be utilized to enable the planning of robotic inspection missions.  
This work is motivated by the need for automated inspection in many commercial and industrial settings, where mobile robots have the navigation capabilities to actually execute the missions, but where planning the path that the robot should take is still a cumbersome process.
In these large, dynamic environments, and in unconstrained spaces such as disaster zones, it's often not possible to augment the environment with fiducial markers to facilitate the sharing such spatial information from the planner to the robot.
We therefore propose to use shared, world-locked coordinate systems through mixed reality as a way to provide a common spatial reference for the user and the robot. 

Existing commercial solutions for mission planning either use a computer interface that is decoupled from the environment, or require the user to drive the robot through the trajectory first, in a teach-and-repeat fashion.
Planning a robot trajectory within a high-fidelity 3D model or mesh of the environment using a computer is certainly possible, and is currently deployed in commercial services (e.g. ROCOS\footnote{https://www.rocos.io/content/robotics-fleet-management}).
However, this approach requires the model to contain a high level of detail of the inspection targets, in order to accurately indicate where the robot should observe without actually being there.
This type of approach has been used for ground robot~\cite{gianni2013augmented} and drone mission planning in mixed~\cite{vaquero2020technologies} and virtual reality~\cite{liu2018usability}, but using 3D terrain models, which are far more available than high-resolution digital twins of inspection sites for ground robots.
Additionally, augmented~\cite{fang2009robot} and mixed reality~\cite{ostanin2018interactive} have been used with a virtual robot arm to facilitate trajectory programming, but connecting these simulated arms to the real robot has used fiducial markers, rather than colocalization with any spatial context.

Unlike these prior works, utilizing the spatial context of the environment provides a more natural way to define a robot trajectory.
However, while solutions such as Boston Dynamics' Autowalk feature for SPOT enable the user to plan missions in context while physically present in the space, such a teach-and-repeat approach requires the user to manually control the robot through the environment each time they define a mission, and it is not possible at this time to edit missions without re-recording them.
While this is a powerful feature, it still has a strict requirement that both the user and the robot are present and available each time a mission must be defined.
The inability to change saved missions also leads to a situation where missions should be executed repeatedly in order for the process of defining them to be worth the effort, otherwise the robot could just be used in teleoperation mode with the same level of efficiency.

\subsection{Mission Planning}
\label{sec:mp_results}
\begin{figure}
    \centering
    \begin{subfigure}[b]{0.45\textwidth}
        \centering
        \includegraphics[width=\textwidth]{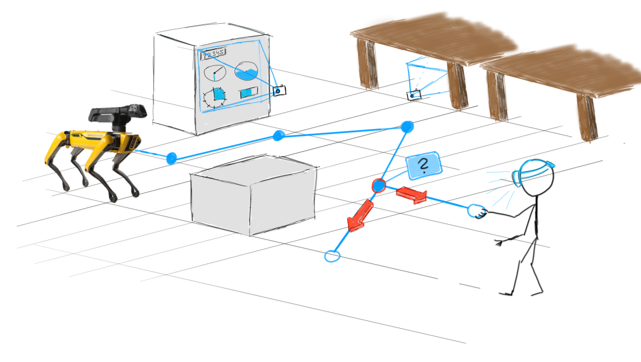}
        \caption{}
        \label{fig:mp_concept_1}
    \end{subfigure}
    \hfill
    \begin{subfigure}[b]{0.45\textwidth}
        \centering
        \includegraphics[width=\textwidth]{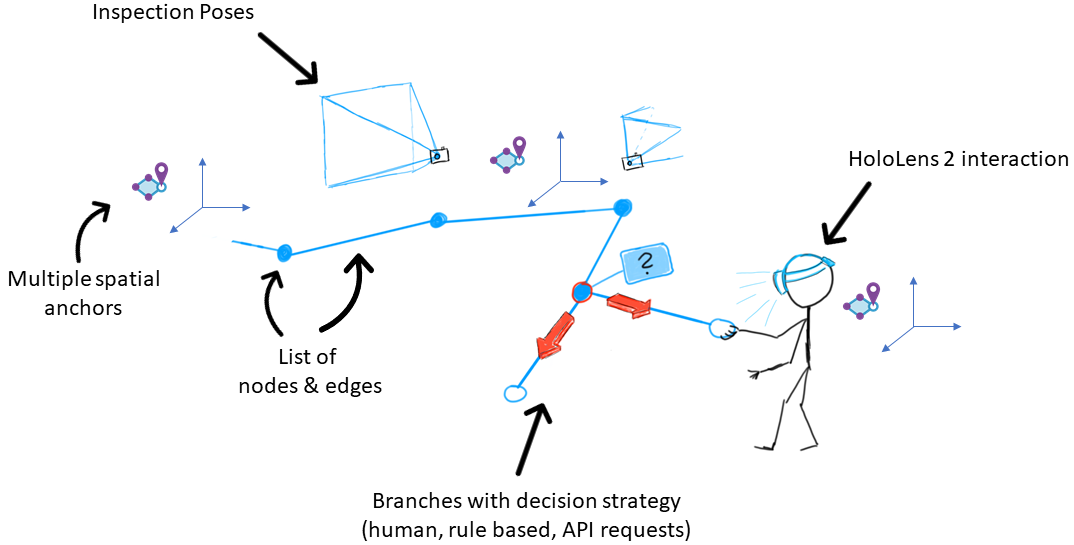}
        \caption{}
        \label{fig:mp_concept_2}
    \end{subfigure}
    \caption{Artistic renderings of mission planning workflow and mission components. A human user with a HoloLens device moves through the environment to be inspected and places holograms that represent waypoints defining a trajectory.  Inspection poses can be defined to trigger the robot to capture data at regions of interest. The underlying structure of a mission is a list of nodes and edges representing a graph of waypoints, with branching decisions in the graph handled by one of several decision strategies.  As the user moves through the space and places waypoints, Azure Spatial Anchors are placed automatically to cover the trajectory so each robot pose can be defined with respect to a nearby reference coordinate system.  This mission structure is then serialized in JSON format, so that it can be retrieved from a database by the robot and executed autonomously after localizing to the spatial anchors in the mission.}
    \label{fig:mp_concept}
\end{figure}

We propose instead to use mixed reality as a tool for defining inspection missions in context, without the requirement of first teleoperating the robot through the desired trajectory.
This concept is illustrated in Fig.~\ref{fig:mp_concept_1}.
This is made possible by using a common coordinate system that persists in space, over time, to share spatial information about the mission between devices.
In particular, we leverage the cloud-based localization service \textit{Azure Spatial Anchors} to create such a reference coordinate frame, and then to enable localization to it by other devices, and persistence of the spatial data.
In this system, a human user with a HoloLens defines an inspection mission by placing holographic markers to indicate desired poses for the robot.
At some later time, an autonomous robot can localize to the spatial anchors that were placed during mission creation, obtain the set of waypoints and inspection poses of the mission, which are defined relative to these anchors, and then execute the mission autonomously.
This approach has the advantage that the robot, whose availability is likely a bottleneck in current work environments, is only required for executing the mission, and not also for defining it.
Furthermore, these missions can also be edited in mixed reality, without the need to recreate an entirely new trajectory each time the mission needs to be adapted.

\begin{figure}
    \centering
    \begin{subfigure}[t][5cm][t]{0.475\textwidth}
        \centering
        \includegraphics[width=\textwidth]{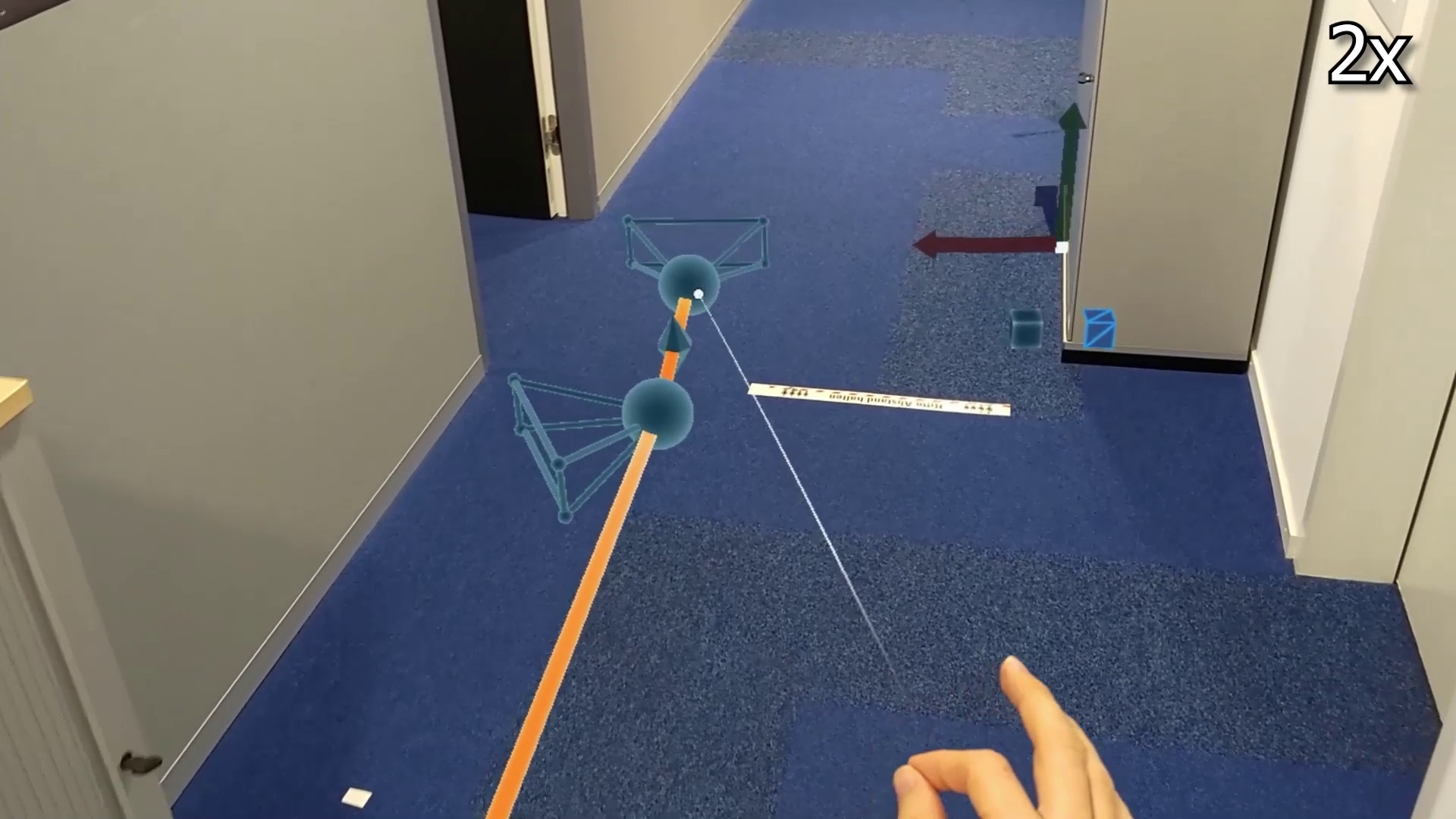}
        \caption{}
        \label{fig:mp_mission_def}
    \end{subfigure}
    \hfill
    \begin{subfigure}[t][5cm][t]{0.475\textwidth}
        \centering
        \includegraphics[width=\textwidth]{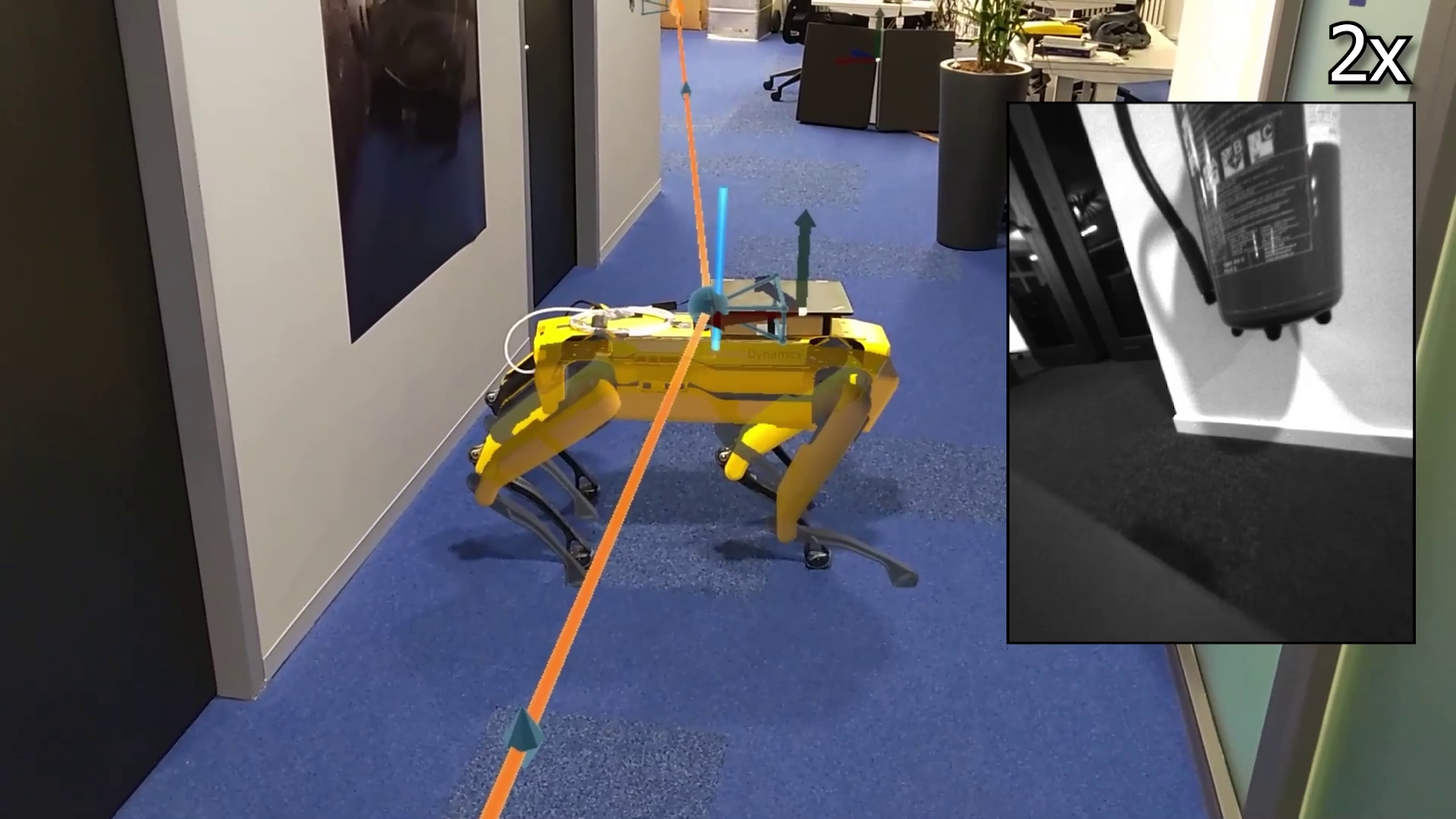}
        \caption{}
        \label{fig:mp_mission_exec}
    \end{subfigure}
    \vspace{0.3cm}
    \caption{Mission planning and execution. Fig.~\ref{fig:mp_mission_def} shows the view through the HoloLens as the user moved holographic markers for the trajectory waypoints. Each waypoint is represented by a sphere for the desired position, and a camera frustum to define the orientation the robot should take for inspection. The axes to the right of the trajectory mark the origin of an automatically-created spatial anchor coordinate system. Fig.~\ref{fig:mp_mission_exec} shows the user's view as the robot is executing the mission, with a robot model overlaid on the physical one, and a sample data capture at an inspection pose. This user view during mission execution is only for illustration, as the robot can execute the missions autonomously without human intervention.}
    \label{fig:mp_missions}
\end{figure}

In practice, the workflow for this system proceeds as follows.
The HoloLens user moves around her environment, manipulating holograms representing inspection waypoints.
These waypoints can be created, deleted, and connected in an arbitrary graph structure, including branching.
Since localization to spatial anchors depends on observing the same part of the environment where the anchor was created, anchors are created automatically as the user moves away from existing anchors, ensuring that waypoints are always defined with respect to an anchor to which they are in close proximity.
We use a radius of $2.5$m from previous anchors as the threshold for creating a new anchor, which was empirically determined.
Anchor localization accuracy and recall degrade beyond $4 - 5$m, but an appropriate value here would depend on the structure and appearance of the environment.
Once the user is satisfied with the mission waypoints, the mission can be saved by serializing it and storing these parameters in a cloud-based database.
The user experience during mission planning is shown in Fig.~\ref{fig:mp_mission_def}, where the waypoints are represented by spheres with camera frustums for the orientation of the inspection pose, and coordinate axes representing spatial anchor reference frames.

When the robot is ready to execute the mission, it obtains the serialized parameters from the database, localizes to the spatial anchors, and then proceeds through the waypoints.
If the HoloLens user is colocalized to the anchors and running the app, the user can monitor the progress of the mission in mixed reality, with an articulated model of the robot overlaid on its pose in the real world (see Fig.~\ref{fig:mp_mission_exec}).
Branching decisions at interconnected waypoints can be made through API calls based on the results of the inspection, or by the user selecting the desired branch if the mission is being executed in this interactive mode with the HoloLens.
The structure of a mission is conceptualized in Fig.~\ref{fig:mp_concept_2}.

\subsection{Azure Spatial Anchors}
\label{sec:asa}
Azure Spatial Anchors (ASA) is a mixed reality cloud service designed for localization.
The fundamental concept of the service is that a small visual map, created through structure-from-motion and representing a specific location in the world, is stored in the cloud with a unique ID.
The set of features and descriptors constituting this map then define a coordinate system that is fixed to the world---if those features can be detected later, and their 3D positions recovered, then the coordinate system can also be recovered.
Devices wishing to localize to a particular anchor, known by its unique ID, then create their own structure-from-motion map, which they submit to the cloud service as a query.
If correspondences for the feature points in the query can be found in the reference map in the cloud, then the service returns a relative pose.

This service is cross-platform and relies on a device's onboard visual odometry in order to build these visual maps to create or query an anchor. 
On HoloLens devices, the onboard head-tracking processes handle this.
For mobile devices, ARKit and ARCore provide the image features and poses for iOS and Android, respectively.
On robotic systems, a special research version of the SDK was released with Ubuntu Linux support as well as a ROS wrapper.\footnote{https://github.com/microsoft/azure\_spatial\_anchors\_ros}
Unlike the other devices, where the visual pose estimation happens in the context of a platform-specific odometry system (e.g. an ARCore or ARKit session), pose estimation solutions on mobile robots can be quite diverse.
The Linux version of the SDK thus accepts undistorted images and poses and computed its own features. 
The pose of the camera can be estimated directly through visual odometry on the robot, or by attaching the camera with a calibrated transformation to a mobile base that is estimating its pose through some other means (e.g. with LiDAR).
While the success of localization queries to ASA depends heavily on the particular cameras, trajectories, and environments in question, in practice the performance is in the range of centimeter-level accuracy. 
This is shown in Fig.~\ref{fig:mp_mission_exec}, where the robot's estimate of its pose with respect to the anchor closely matches the actual robot pose when it is transformed into the HoloLens' field of view, as seen in the close alignment of the robot and its holographic model.
\subsection{System Overview}
\label{sec:mp_system}
\begin{figure}
    \centering
    \includegraphics[width=\linewidth]{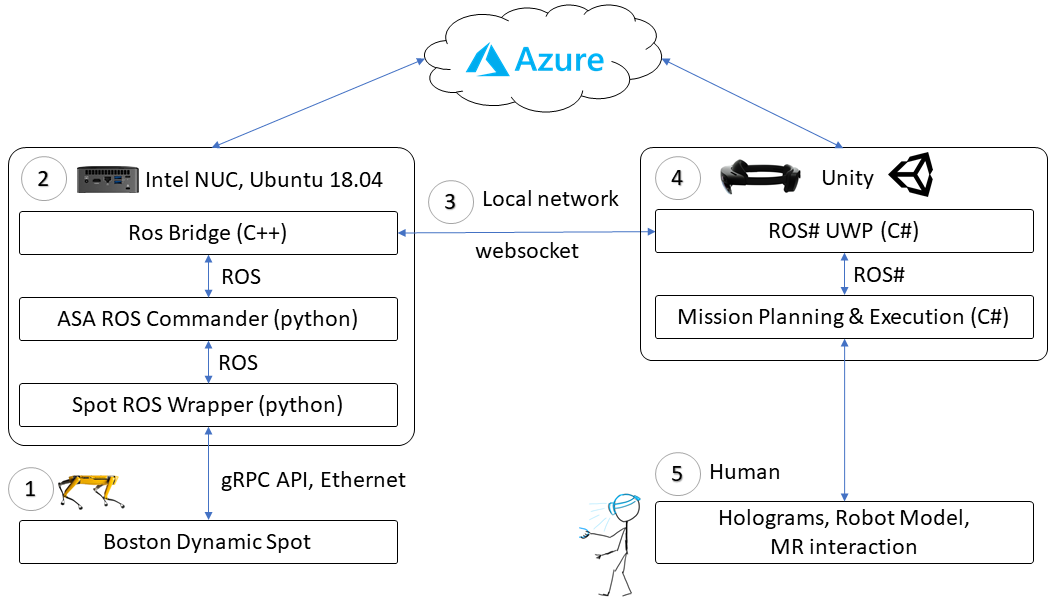}
    \caption{System diagram showing the components of the mission planning framework.  The robot (1) communicates through its SDK to a ROS wrapper and our mission planning orchestration node, which run on a companion computer (2).  This computer communicates with the HoloLens over the local network (3) or through serialized missions stored in an Azure database. The HoloLens (4) interfaces with the ROS communication framework using a ROS\# Unity plugin to the mission planning application, which ultimately provides an interactive, mixed reality experience for the user (5).}
    \label{fig:mp_system_diagram}
\end{figure}
The system is organized in a modular and distributed fashion, with components deployed on the HoloLens, on the robot, and in the cloud.
It is possible to run the HoloLens and robot at the same time, where the mission parameters and some visualizations are sent between the devices with ROS, or to define and run the mission at different times, in which case the mission definition is serialized and stored in a databse in the Azure cloud.
An overview of these components and the flow of data between them is visualized in Fig.~\ref{fig:mp_system_diagram}.

The HoloLens runs an application that is built using the Unity 3D engine\footnote{https://unity.com/}.
Some pre-built components from the Mixed Reality Toolkit\footnote{https://github.com/microsoft/MixedRealityToolkit-Unity} handle the user's interaction with the app's holograms.
The ROS\#\footnote{https://github.com/siemens/ros-sharp} package acts as a bridge between the Unity environment and the ROS communication framework.
HoloLens provides support for creating and querying Azure Spatial Anchors through an SDK, and this functionality is connected to Unity through an XR plugin layer.
These components work together to enable the user to create and move holographic markers in the environment, and store the poses of these waypoint markers with respect to spatial anchor coordinate systems.

On our robot, a Boston Dynamics SPOT, we connect a companion computer (an Intel NUC running Ubuntu 18.04) through the robot's ethernet port, in order to run our user code.
This code includes of a wrapper for SPOT's gRPC-based SDK\footnote{https://github.com/boston-dynamics/spot-sdk} that connects its data streams to ROS topics\footnote{https://github.com/clearpathrobotics/spot\_ros}, as well as a corresponding bridge node that receives messages from the ROS\# component in the HoloLens Unity app that are sent over a websocket.
An additional ROS package of utilities\footnote{https://github.com/EricVoll/spot-mr-core} helps to interface SPOT with the Azure Spatial Anchors service.
The robot's behavior is orchestrated by the \textit{ASA ROS Commander} node, which obtains the mission definition from either ROS or the cloud database.
Once the robot localizes to the spatial anchors defined in the mission using the ASA service, this node sends the waypoints as trajectory commands through the wrapper to the Spot SDK, and monitors the robot's progress through the mission.
An important note is that this mission planning system relies on the robot's underlying navigation capabilities to actually plan and execute paths to the waypoints that are sent as goals.

Finally, the cloud components consist of the Azure Spatial Anchors service (see Sec.~\ref{sec:asa}), and a CosmosDB database to store serialized mission parameters.
The waypoint poses, spatial anchor IDs, and other mission information are serialized into JSON format and then stored in the database with a unique ID as a key.
When operating in the temporally decoupled mode, the \textit{ASA ROS Commander} node can look up a particular mission in the database, download the JSON, and then execute the mission once it localized to the mission's anchors.

\subsection{Outlook}
This system represents a milestone in spatial computing and interaction. 
By sharing common reference frames, the mixed reality device can provide the robot with actionable spatial information that it can understand in its own spatial context, leading to an improvement in efficiency in a common commercial and industrial robotics task.
One limitation of the current spatial anchors service is that the anchor maps are not automatically connected together in the cloud to form large-scale, continuous digital twins, even if they are located in the same area.
This use of discrete spatial anchors limits scalability for this application, but future work on using continuous digital twins for localization of multiple heterogeneous devices will enable this type of shared spatial understanding on a large scale.

%% file: colocalization.tex
\section{Colocalization and Interaction}
\label{sec:colocalization}
Colocalization of devices requires that they are each able to localize themselves to a common reference coordinate system.  
Through their individual poses with respect to this common coordinate frame, the relative transformation between localized devices can be computed, and subsequently used to enable new behaviors and collaboration between devices.
In the scenario described in Sec.~\ref{sec:mission_planning}, the robot and mixed reality device do not interact directly and are not necessarily colocalized at the same time, but share spatial information that is anchored to a common spatial reference frame.
Here, we consider a scenario in which multiple devices colocalize to a shared coordinate system simultaneously, thereby enabling temporally synchronized behaviors and interaction.

When the colocalized devices are all mobile robots, this could mean that swarming or collective behaviors are possible, as well as parallelization of spatial tasks.
On the other hand, if there is a mix of human-centered devices (e.g. mixed reality headsets or mobile devices) and robots, this colocalization can unlock more natural human-robot interaction.
This section describes a system that exploits the colocalization of a human wearing a Microsoft HoloLens 2 and a mobile robot to demonstrate intuitive hand gestures for robot control. 

This work is motivated by a need for semi-autonomous behaviors with shared control, as a way to reduce the attentional load for the operator.
Defining high-level tasks for the robot to perform autonomously, particularly through intuitive interfaces, enables the user to focus on other tasks or control multiple robots simultaneously.
This is desirable in search and rescue environments~\cite{delmerico2019current}, but will be important in the increasingly robot-filled workplace of the future.

\subsection{Colocalizing the Human and Robot}
\label{sec:colocalizing_human_robot}
\begin{figure}
    \centering
    \begin{subfigure}[t][5cm][t]{0.475\textwidth}
        \centering
        \includegraphics[height=5cm]{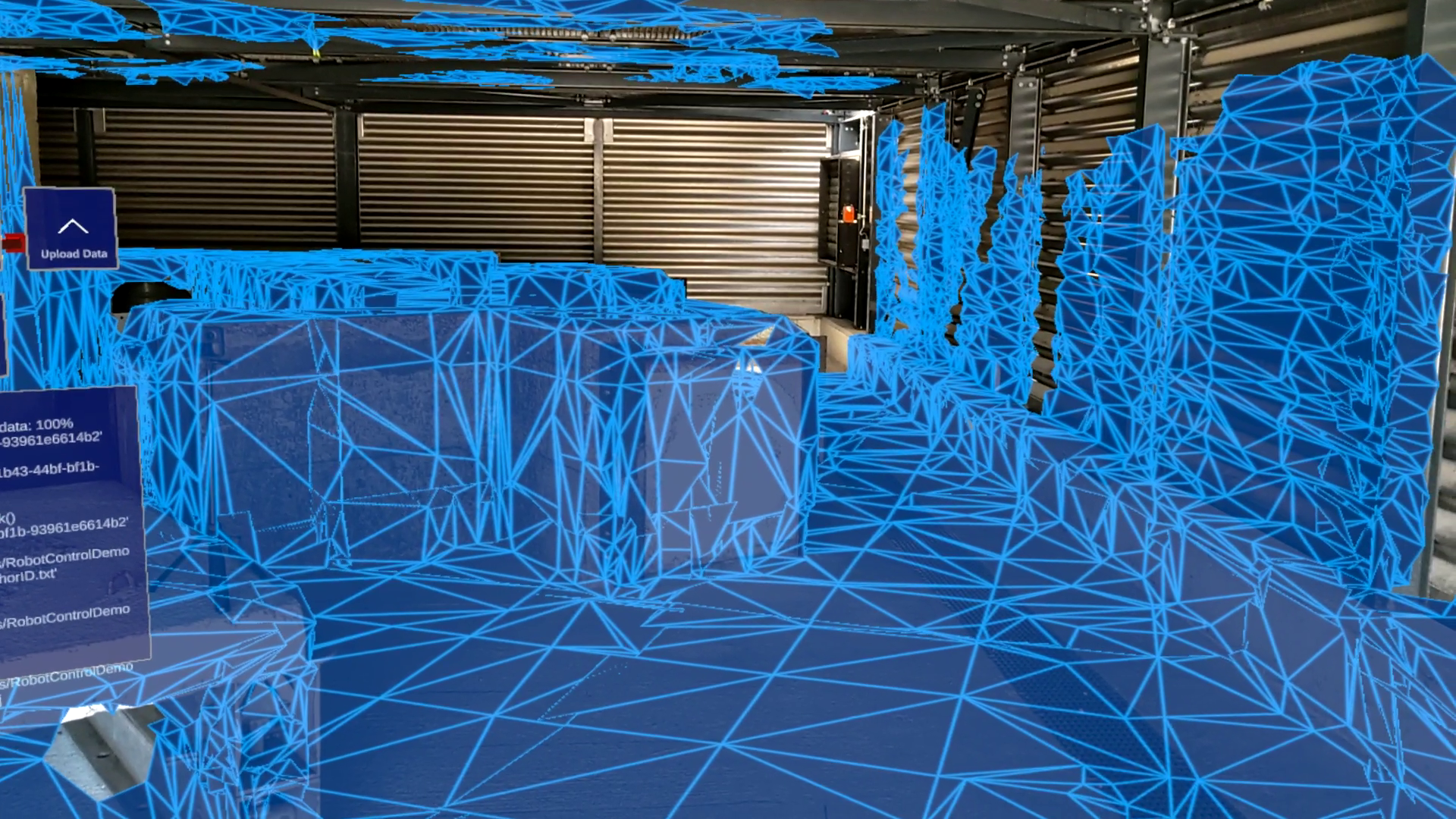}
        \caption{HoloLens view}
        \label{fig:shared_map_hololens}
    \end{subfigure}
    \hfill
    \begin{subfigure}[t][5cm][t]{0.475\textwidth}
        \centering
        \includegraphics[height=5cm]{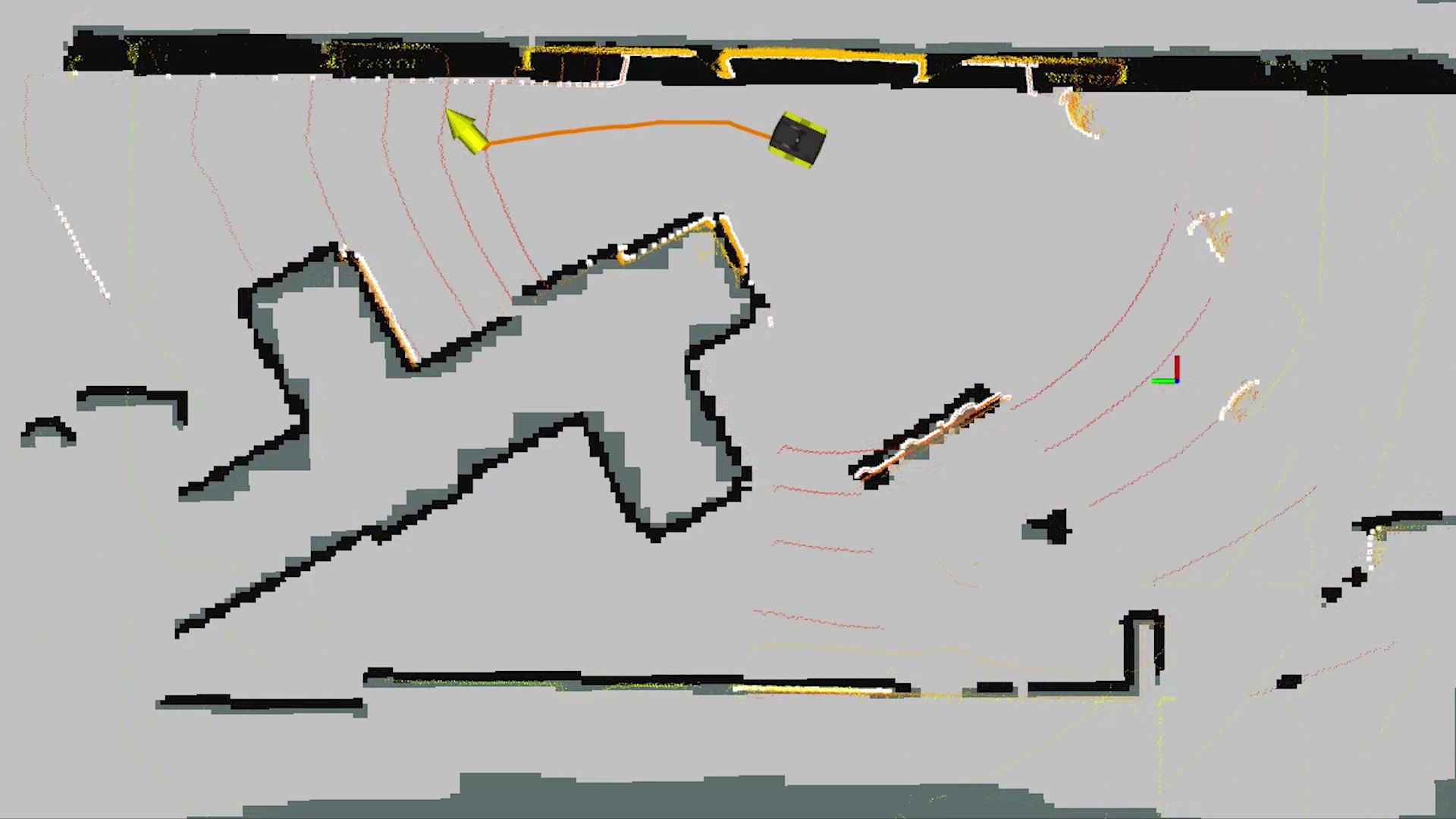}
        \caption{Robot view}
        \label{fig:shared_map_robot}
    \end{subfigure}
    \vspace{0.4cm}
    \caption{Colocalization of a robot and a HoloLens through a shared map. Fig.~\ref{fig:shared_map_hololens} shows the user's view of a spatial mesh, which was captured from the HoloLens, overlaid on the real world. This map is converted to a 2D occupancy grid representation, whose coordinate frame is aligned with that of the mesh, in order to enable robot localization with LiDAR (shown in Fig.~\ref{fig:shared_map_robot}).}
    \label{fig:shared_map}
\end{figure}
We consider two approaches for colocalizing devices to a common reference frame: sharing a map, and utilizing a visual localization service in the cloud.
Prior work in this domain has explored the use of augmented reality (AR) for more efficient human-robot interaction and visualization of the robot's state and intent.
However, these works have relied on colocalization through detection of landmark objects~\cite{chandan2021arroch}, or the use of fiducial markers mounted on the robot~\cite{muhammad2019creating}.
While these approaches for colocalization are sufficient to perform AR visualization, they do not provide further spatial understanding for the AR device or robot.
Here, we propose to share not just a relative pose but a shared map between the devices, enabling both devices to take advantage of a common digital twin of the space.

In order to share a map from the HoloLens with the robot, we execute an offline procedure to create and then convert a map of an environment for colocalization.
We leverage the onboard spatial mapping processes of the HoloLens, which constantly build a sparse visual feature map for tracking the motion of the device, as well as a dense mesh of the environment.
The user observes the environment with the HoloLens depth camera in order to build the dense mesh, and our application provides visual feedback to show areas of the space that have been mapped (see Fig.~\ref{fig:shared_map_hololens}).
The sparse map is aligned to the dense representation, and it enables the HoloLens to relocalize to the environment in a future session.

Once the spatial mesh has been captured, we apply several processing steps to convert it to an occupancy grid representation that can be used by the robot for LiDAR-based localization.
We take the mesh as input, which typically consists of several connected components, and apply Poisson reconstruction to make it watertight.
This watertight mesh is used to initialize a signed distance function (SDF) representation of the space.
Finally, a horizontal 2D slice from the SDF is extracted at a user-defined height, such that the implicit surfaces in the SDF define the occupied cells in the occupancy grid representation, and the voxels in the SDF with positive distance represent free space cells.  
An example of such a map being used for LiDAR localization is shown in Fig.~\ref{fig:shared_map_robot}.

The X-Y origin of the map is preserved during these conversion steps, so 2D coordinates given in the HoloLens coordinate frame correspond to the same X-Y position in the plane where the robot is navigating.
So in the shared map scenario, when the HoloLens and the robot localize to their respective maps, spatial information can be translated between the two map representations.

We also demonstrate the use of the Azure Spatial Anchors (ASA) cloud localization service to colocalize the HoloLens and robot to the same anchor. 
This service was described in more detail in Sec.~\ref{sec:asa}. 
For this application, we require a single anchor that both devices can localize to, which then provides a common reference frame.
In our workflow, the HoloLens observes the environment and then creates an anchor through the service.
We synchronize the unique ID of the created anchor with the robot through ROS, and then leverage the ASA ROS wrapper to query this anchor using a stream of camera images and poses from the robot.
Once the robot localizes to the desired anchor, the anchor's reference frame is added to the ROS \textit{tf} tree, where it can be used to transform spatial data to or from the robot's other reference frames.

For the purposes of our system, these two approaches are interchangeable.
They both provide the devices with a common coordinate system through which to share spatial information, but the shared map approach is tailored to robots equipped for LiDAR navigation, while the ASA approach is designed for robots using visual navigation.
\begin{figure}
    \centering
    \includegraphics[width=\linewidth]{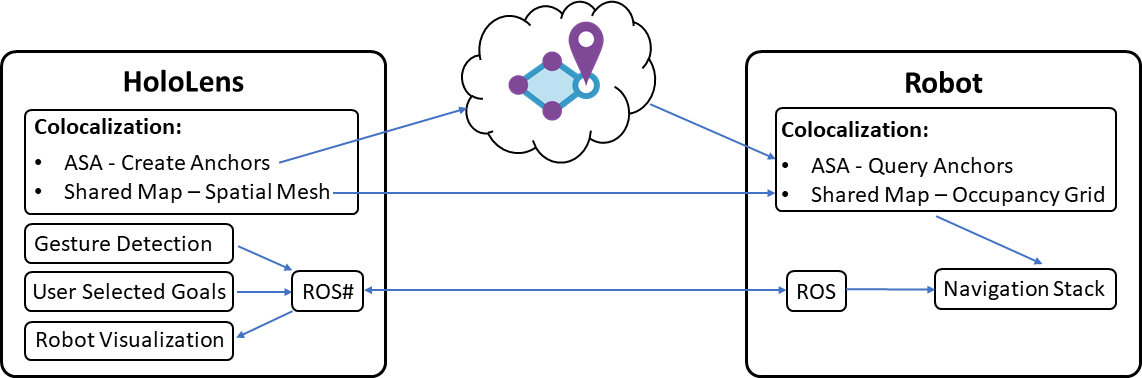}
    \caption{System diagram for the interaction framework. Colocalization between the HoloLens and robot is achieved either by both devices localizing to the same Azure Spatial Anchors, or by creating a map on the HoloLens (a 3D mesh) and sharing it with the robot in a representation that it can use (an occupancy grid).  The devices communicate spatial information (navigation goals, robot states and paths), which is defined with respect to their shared coordinate system, via ROS.  Human-robot interaction is achieved through intuitive gesture recognition, or by placing holographic markers to give target poses to the robot.}
    \label{fig:colocalization_system_diagram}
\end{figure}

\subsection{Gesture Recognition}
\label{sec:gesture_recognition}
In the colocalized scenario we consider here, egocentric sensing by human-oriented devices can give spatial meaning to human motions, expressions, and gestures.
We focus on detecting and classifying hand gestures with a HoloLens 2 headset, and then translating these into navigation commands for the robot.
The HoloLens 2 features a variety of cameras for sensing both the environment and the actions of the user, and estimating how the device moves through space.
In particular, we leverage its hand tracking capability, which uses the depth camera to track the user's articulated hands and exposes the joint positions through an API.  

We implement our gesture classification model as a neural network.
Namely, we use a MultiLayer Perceptron (MLP)~\cite{goodfellow2016,hastie2009} which takes as input HoloLens hand tracking data and outputs probabilities over a set of predefined gesture classes.
As for the hand data, we assume that only one hand is in view (or we consider the right hand, if both are in view), and extract local joint angles for a total of 19 joints.
In our experiments, we found that using only joint \emph{flexion} angle values is enough to obtain an accurate gesture classification. We therefore parameterize each joint by a single value. Relying on single-frame predictions can lead to noisy classification results.
To improve robustness we leverage temporal information, running the model over time windows of 12 frames (which correspond to roughly $0.2$ seconds with the app running at 60fps): to classify a gesture at time $t$, we consider all the hand joint angle values tracked within the interval $[t, t-12)$, flattening them into a $(12 \times 19)$-dimensional vector which constitutes our MLP input. 
For the output, we consider three main gesture classes for human-machine interaction: ``stop'', ``come here'' and ``point''. We also add a background class to identify frames in which the user is not interacting with the robot and therefore not performing any gesture. This gives us a total of 4 classes.

The MLP network has a total of 4 layers and uses Rectified Linear Units (ReLUs)~\cite{goodfellow2016} as activation functions. All hidden layers are $128$-dimensional. Per-gesture confidence values are obtained by applying the sigmoid function to the output of the last layer. 
Finally, the gesture with the highest confidence is chosen as the classifier output. Figure~\ref{fig:gesture_classifier} exemplifies the app output: the first line describes the output gesture (``stop''), which is the one obtaining the highest confidence value ($0.62$).
We found that adding an attention layer~\cite{vaswani2017} right before the classification one helps capture spatial and temporal correlations between joints, and therefore yields more accurate results.

The network showed good results already when trained on a small training set: we asked 6 subjects to perform each of the 3 gestures (plus random actions for the background class) twice, recording the corresponding hand tracking data with an ad-hoc app. We trained our model on data from 5 subjects, withholding 1 subject for validation.  
We implemented our MLP network in pytorch, and then reimplemented it in C\# to perform inference in Unity.

\subsection{Robot Control with Holograms and Gestures}
\begin{figure}
    \centering
    \begin{subfigure}[t][5cm][t]{0.27\textwidth}
        \centering
        \includegraphics[height=5cm]{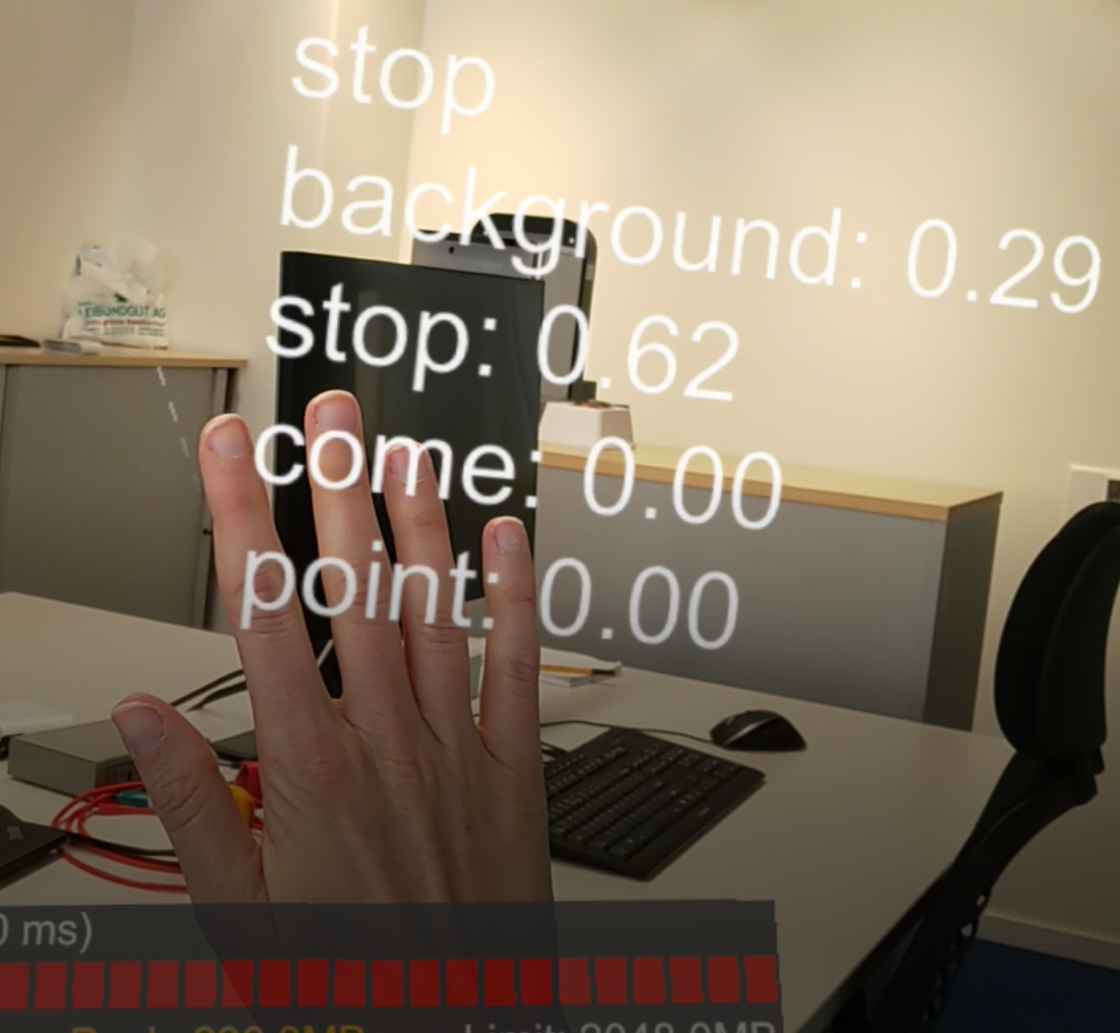}
        \caption{Gesture classifier output}
        \label{fig:gesture_classifier}
    \end{subfigure}
    \hfill
    \begin{subfigure}[t][5cm][t]{0.24\textwidth}
        \centering
        \includegraphics[height=5cm]{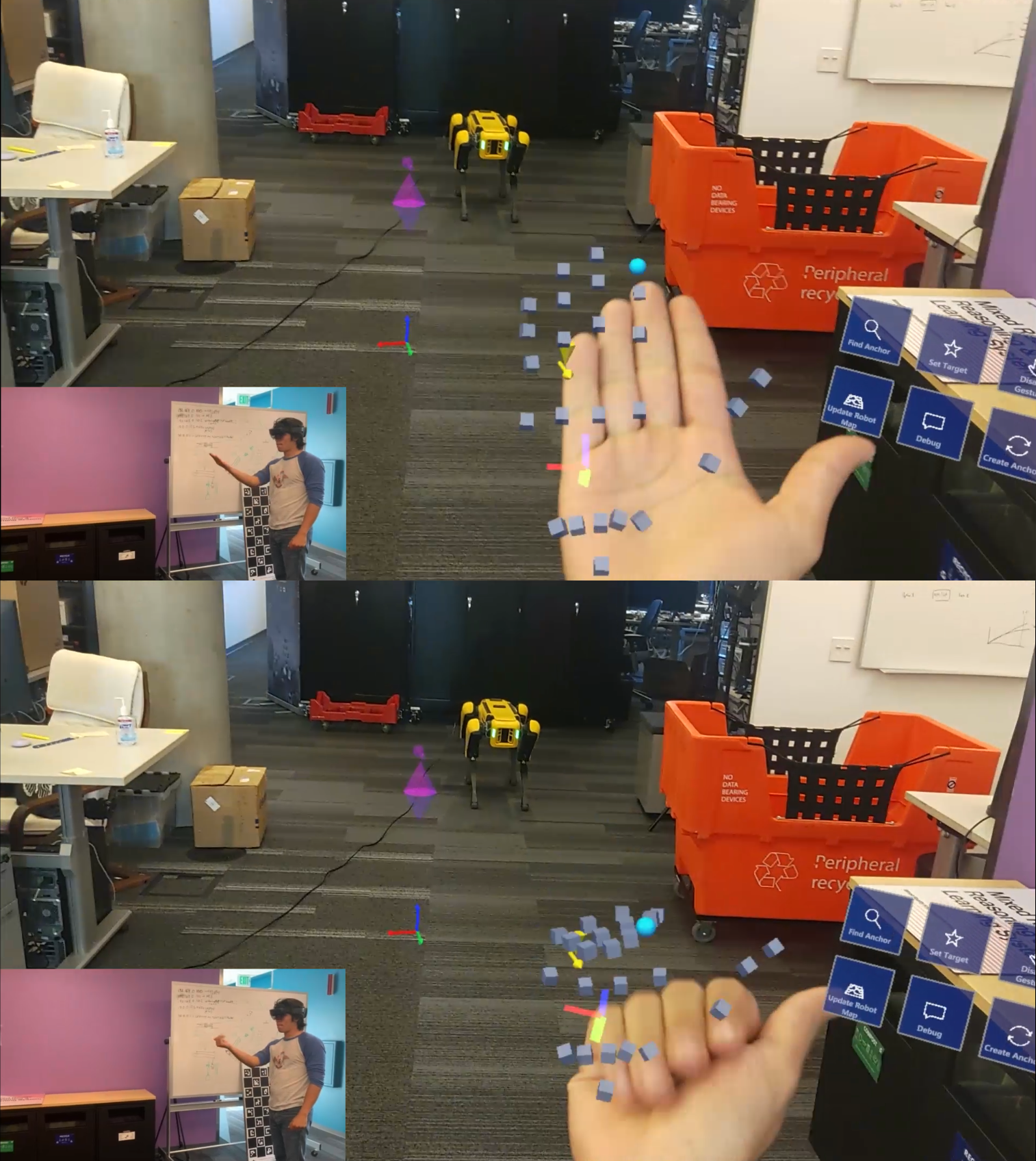}
        \caption{``Come here'' gesture}
        \label{fig:come_here_gesture}
    \end{subfigure}
    \hfill
    \begin{subfigure}[t][5cm][t]{0.475\textwidth}
        \centering
        \includegraphics[height=5cm]{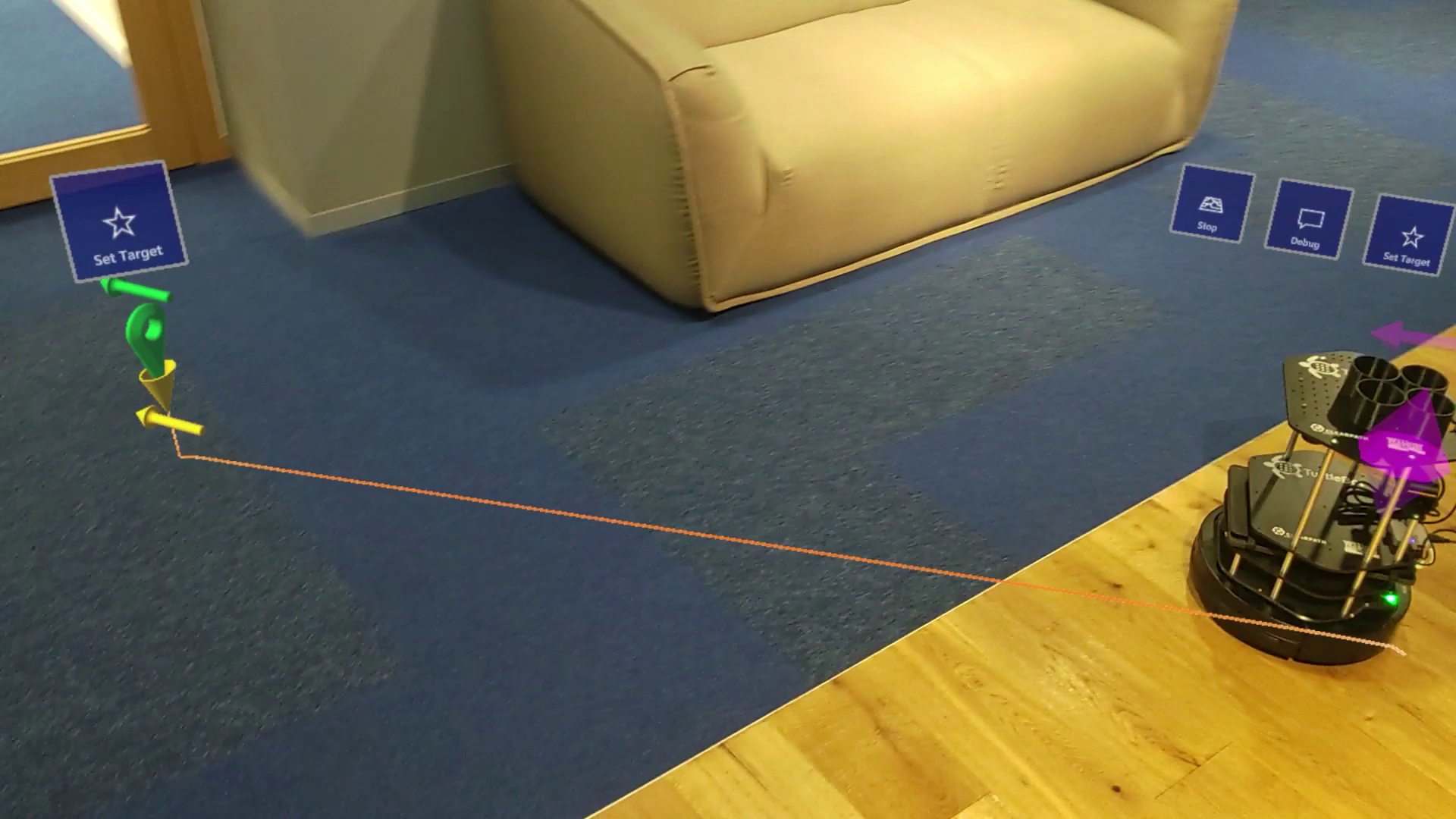}
        \caption{Planned path showing robot intent}
        \label{fig:planned_path}
    \end{subfigure}
    \vspace{0.2cm}
    \caption{Human-robot interaction through hand gestures and navigation goals.  In Fig.~\ref{fig:gesture_classifier}, the user is making the ``stop'' gesture, which the classifier correctly identifies. This gesture is mapped to an action that preempts and cancels an existing trajectory on the robot.  Another gesture that is recognized is ``come here'', shown in Fig.~\ref{fig:come_here_gesture}, which triggers the robot to plan a path to the position of the user, which it knows due to colocalization.  Finally, the robot can provide visual feedback of its intent, as seen by its planned path in Fig.~\ref{fig:planned_path}.}
    \label{fig:gestures}
\end{figure}

In this system, the HoloLens acts as a human-robot interface for giving the robot navigation commands.
The user can select an arbitrary navigation goal for the robot by moving a holographic marker to a location in the environment and confirming this as the robot's target pose by clicking a holographic button (see Fig.~\ref{fig:planned_path}).
Alternatively, intuitive hand gestures (see Sec.~\ref{sec:gesture_recognition}) are detected on the HoloLens and then translated to navigation commands for the robot.
Figure~\ref{fig:colocalization_system_diagram} shows a diagram illustrating the components of the system and the flow of data between them.

In the case of a navigation goal set by the holographic marker, the target pose is selected in the reference frame of the HoloLens, but is transformed to the common reference frame of the map that is shared between the robot and HoloLens, in order for the robot to interpret that goal in its own reference frame.
The \textit{come here} gesture shown in Fig.~\ref{fig:come_here_gesture} leverages the shared spatial understanding of the two devices to provide the semantically meaningful action that corresponds to the gesture.
When the gesture is detected, the ``here'' in \textit{come here} is translated to be the HoloLens' location in the shared map, and thus a goal pose in front of this location is sent to the robot.
The \textit{point} gesture works similarly to \textit{come here}, in the sense that when it is detected, the pose of the HoloLens and the user's hand position while making the gesture are used to determine the position that they are pointing to, and this is sent as the goal.
The \textit{stop} gesture is used to preempt execution of the current trajectory if the robot is already moving toward a target pose.  

We use ROS for communication between the HoloLens and robot, with ROS\#\footnote{https://github.com/siemens/ros-sharp} providing the interface between the Unity-based holographic app running on the HoloLens and the other ROS nodes.
The design of the system is robot-agnostic, and relies on the robot's own navigation stack to actually execute any motion commands that are generated from the HoloLens. 
For example, we have demonstrated this system as a human-robot interface for controlling a Turtlebot 2 and a Clearpath Jackal by sending \texttt{PoseStamped} messages from the HoloLens to the robot's \textit{move\_base} node (see Fig.~\ref{fig:planned_path}).
We have also controlled a Boston Dynamics SPOT with the same system, by passing the target pose for the robot to the SPOT SDK's\footnote{https://github.com/boston-dynamics/spot-sdk} \texttt{trajectory\_command} function and allowing its onboard navigation to move the robot to reach the goal. 

An important aspect of using a mixed reality human-robot interface is that spatial information can be shared in both directions---from user to robot, and from robot to user---within the environmental context.
In the case of this system, spatial information shared by the robot takes the form of a marker for its position and holographic line representing the planned path to reach its current target pose.
The ability to visualize the robot's state and intentions simultaneously with the surrounding environment provides a clear advantage in terms of human-robot safety.  
When the HoloLens user can see the path that the robot is following overlaid on the real world, they can understand the robot's intent more clearly than with colored lights or a 2D interface on a mobile device. 
We envision safer and more efficient human-robot collaboration through this type of feedback, where visualization of the robot's intent can fill in the gaps left by body language and verbal communication that make human-human collaboration more straightforward.

\subsection{Outlook}
We have demonstrated that gesture control for robots can be more intuitive when the human and the robot can share a spatial context, and gestures can have spatial meaning.
This is made possible by the egocentric sensing of the HoloLens, and a shared map representation between the devices.
However, the gesture detection system proposed here only captures a small set of commands.
There are many opportunities to expand on these types of spatially-relevant gestural interactions when the robot and human are colocalized through a shared understanding of their space.

%% file: embodiment.tex
\section{Immersive teleoperation and embodiment}
\label{sec:immersive_teleop}
In the previous section, we discussed gesture based interaction in a shared environment. 
Here, we remove the shared environment and shared understanding of space, and instead explore the projection of the user's actions to a remote robot, and the robot's sense of space back to the user.
We consider several levels of immersion, based on \emph{touching} and manipulating a model of the robot to control it, and the higher-level immersion of \emph{becoming} the robot and mapping the user's motion directly to the robot.
The works described here focus on immersive teleoperation of \emph{robotic arms}, where challenges include the differences in structure and articulation between the human and robot, and the lack of proprioception in teleoperation. 
We present several novel approaches that address these challenges through mixed reality: embodied teleoperation, motion retargeting, and task autocorrection.
In each of these methods, a mixed reality device provides a way to collect multimodal inputs (e.g. hand poses), and provide immersive visual feedback to the user.
All of these approaches are motivated by a desire to improve the ease of use and user experience in controlling a robot to perform manipulation tasks.

\begin{figure}
    \centering
    \begin{subfigure}{0.39\textwidth}
        \centering
        \includegraphics[width=\linewidth]{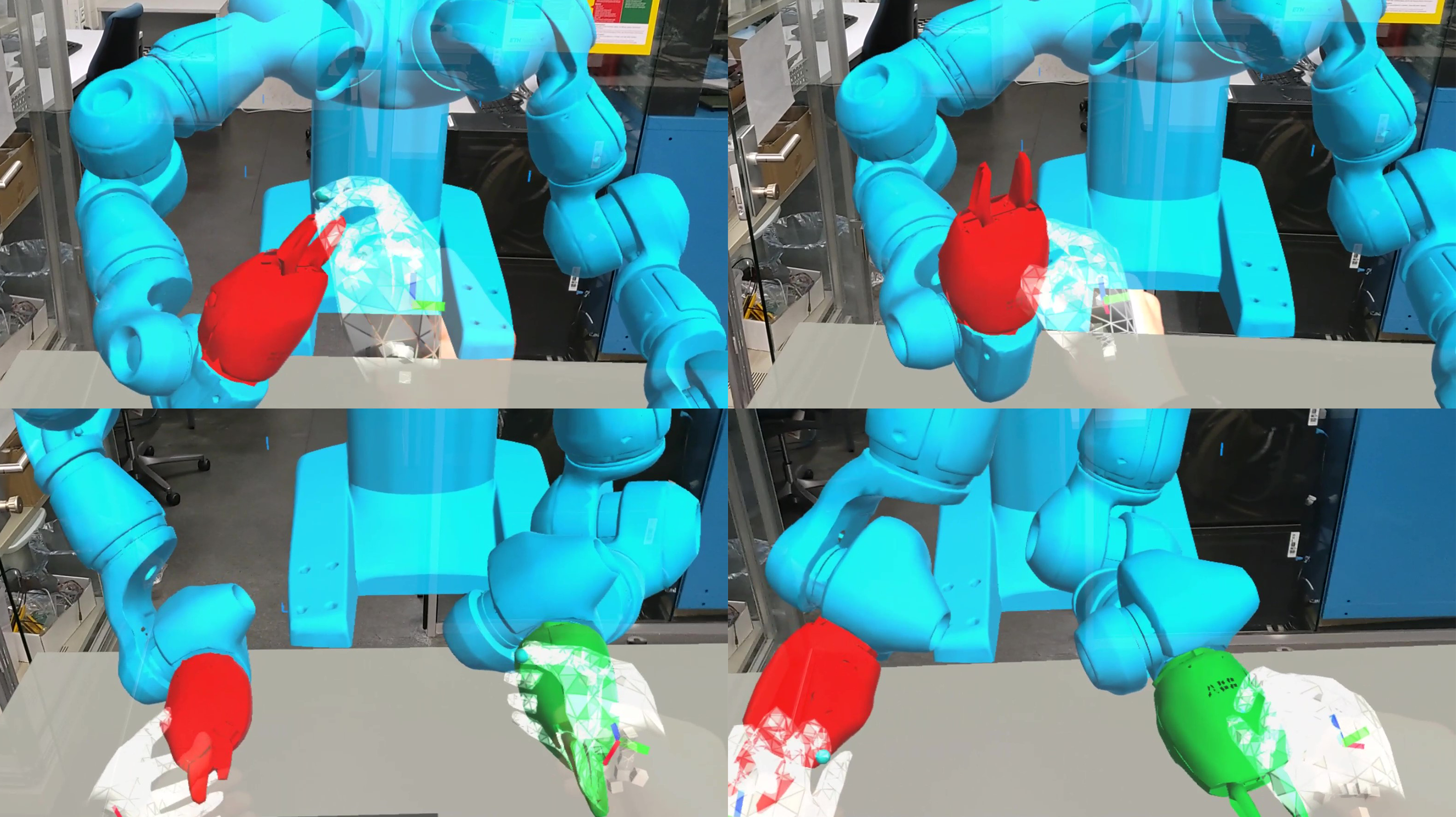}
        \caption{Teleoperation by touch}
        \label{fig:Yumi_touch}
    \end{subfigure}
    \hfill
    \begin{subfigure}{0.59\textwidth}
        \centering
        \includegraphics[width=\linewidth]{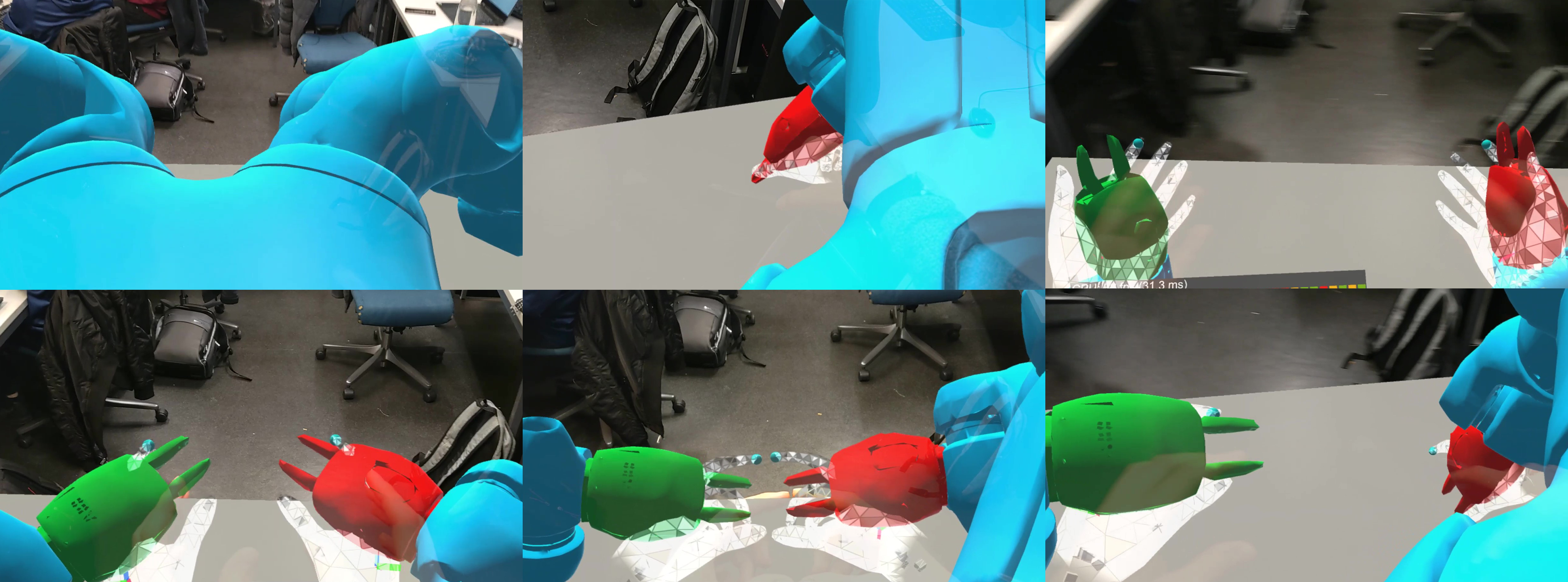}
        \caption{Teleoperation by embodiment}
        \label{fig:Yumi_possess}
    \end{subfigure}
    % \vspace{0.4cm}
    \caption{Teleoperation by controlling a digital twin can be done from the outside, via kinesthetic programming, or from within, by embodying the robot and having it follow the motions of the operator.}
    \label{fig:Yumi_teleop}
\end{figure}

A basic level of proficiency in teleoperating a robotic arm, for example using a joystick, may be reached quite quickly.  
But achieving mastery for performing complicated actions can require a significant time investment, particularly if the arm has many degrees of freedom.
It is mentally taxing, as the operator must internally compute transformations, or rely on muscle memory, to perform even simple manipulation tasks.
One alternative approach for controlling an arm is \emph{kinesthetic programming}, where the operator directly touches and \emph{moves} the arm to record motions and later replay them.
This is impossible to do from afar, but sensory disabilities or limited mobility of the user may also make it impractical to program the desired motions in person.
Instead, immersive teleoperation allows us to manipulate a digital twin of a remote arm by holographic touching (see Fig.~\ref{fig:Yumi_touch}).
% This is demonstrated in \revedit{Fig.}~\ref{fig:Yumi_touch}.
There, using a HoloLens 2, an operator grabs the end effector of Dual-armed ABB YuMi hologram by pinching and repositioning its hologram.
The underlying system simply solves an inverse kinematics (IK) problem in real time, and can stream joint angles to a remote YuMi.

Another emerging approach for immersive teleoperation is via \emph{embodiment}, where the robot is treated as an \emph{avatar} of the operator
In this case, the operator \emph{possesses} the robot by transferring their motions to the robot, and optionally viewing the environment from its own perspective.
An example of this using the same YuMi/HoloLens application appears in Fig. \ref{fig:Yumi_possess}.
In this case, the palms of the users are tracked by the HoloLens, and their positions are set as IK targets for the end effectors of the YuMi.
A pinching gesture also closes the gripper.

A notable demonstration of embodiment that leverages \emph{tactile feedback} is the Tactile Telerobot\footnote{https://www.convergerobotics.com/}, a system that comprises a robot hand with multiple sensors mounted on a robot arm, which is operated using a tactile feedback glove.
Other recent approaches leverage different combinations of tracking methods, HMDs and robotic systems.
For example, a recent vision-based system, DexPilot~\cite{Handa2020}, was used to teleoperate a robotic hand-arm system by observing human hand via 4 RGB-D cameras.

Ultimately, our goal with regards to the embodiment approach is to enhance the sense of \emph{body ownership}. 
That is, the ability of the operator to comfortably control the robot as if it was their own body.
Robots today are still far away from true body ownership, with significantly lower-than-human dexterity, tactile sensing and feedback that can not mimic reality well, and engineering limitations such as high latency and errors in tracking.
We consider some of these fundamental challenges as we explore the mapping of human inputs to robot outputs with \textit{motion retargeting}.

\subsection{Motion retargeting}
\begin{figure}
    \centering
    \begin{subfigure}[t][5cm][t]{0.475\textwidth}
        \centering
        \includegraphics[height=5cm]{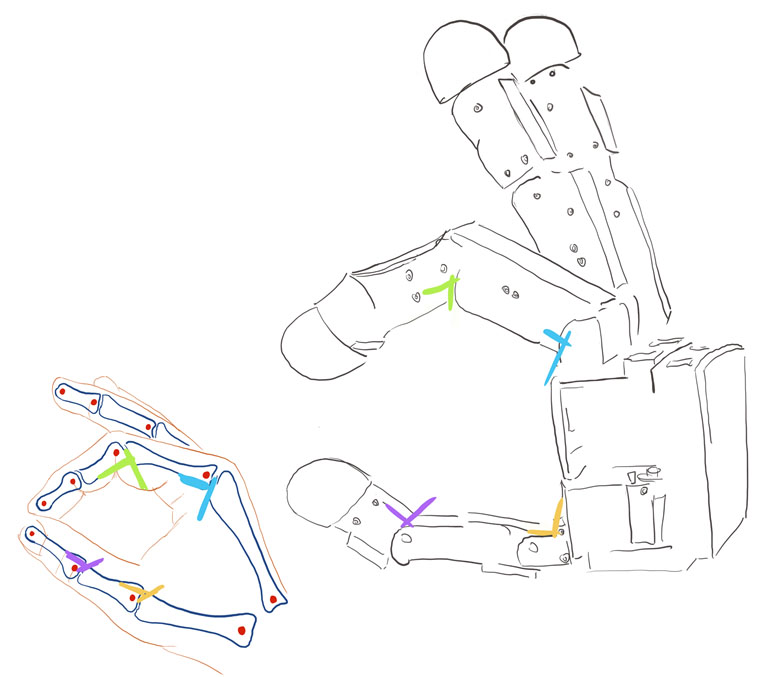}
        \caption{Angle mapping}
        \label{fig:anglemapping}
    \end{subfigure}
    \hfill
    \begin{subfigure}[t][5cm][t]{0.475\textwidth}
        \centering
        \includegraphics[height=5cm]{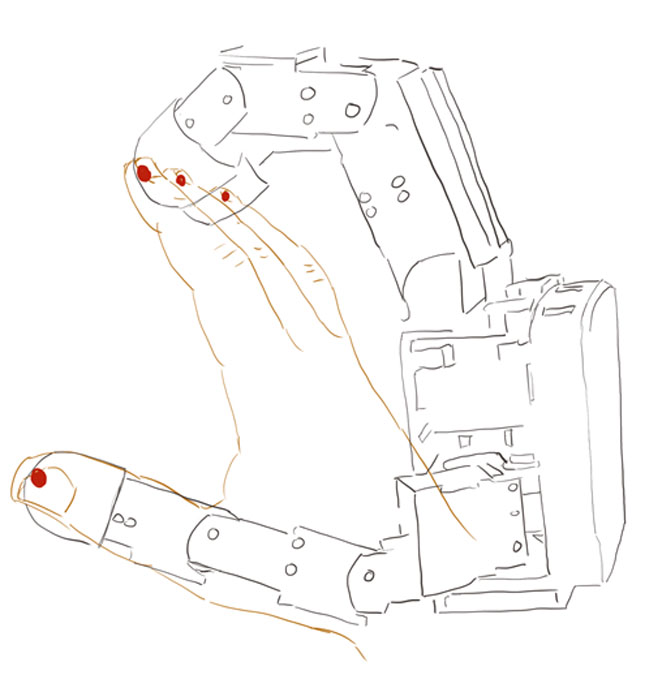}
        \caption{IK-based mapping}
        \label{fig:IKmapping}
    \end{subfigure}
    \vspace{0.4cm}
    \caption{Retargeting hand poses can be accomplished with many different strategies. The simpler strategies involve mapping the operator's joint angle to the robot directly, or using the operator fingertips as inverse kinematics targets. }
    \label{fig:retargeting}
\end{figure}

Central to the question of embodiment is the question of mapping human motions to an avatar.
The difficulty lies in the vast differences between human and robot proportions and morphology.
Two of the most commonly used approaches are joint angle mapping, or inverse kinematics.
Angle mapping maps the operator's joint angles directly to the avatar's joint angles.
Naturally, this is only feasible when the avatar has similar morphology to the human.
While motions mapped with this approach may appear plausible, they lack some \emph{semantics}.
For example, the Fig. \ref{fig:anglemapping} illustrates joint angle mapping of a human hand to a robot hand.
The human hand is posed such that it is touching a finger to a thumb, but this pose is not captured by the robot hand due to these differences in morphology.
Alternatively, the IK approach finds joint angles to map the avatar's hands, fingertips or end effectors, relative to the operator's hand or fingertips (see Fig.~\ref{fig:IKmapping}).
In this case, the operator has more control over the exact positioning of the fingertips, but the hand pose itself might appear very different.
This could be an issue when, for example, the operator wishes the avatar to make a specific hand gesture.

Several alternative approaches for retargeting have been proposed in the past, with the intent of making retargeting more effective for performing tasks, thus also improving body ownership.
These approaches commonly formulate an optimization problem which minimizes various combinations and variations of objectives similar to the ones above.
For example, Rakita et al.~\cite{Rakita2017} used a weighted sum of an IK objective and other smoothness terms, 
while~\cite{Handa2020} minimized a weighted sum of \emph{differences} between the distances of operator fingers and robot fingers, both with heuristically determined weights.

\begin{figure}
    \centering
    \includegraphics[width=\linewidth]{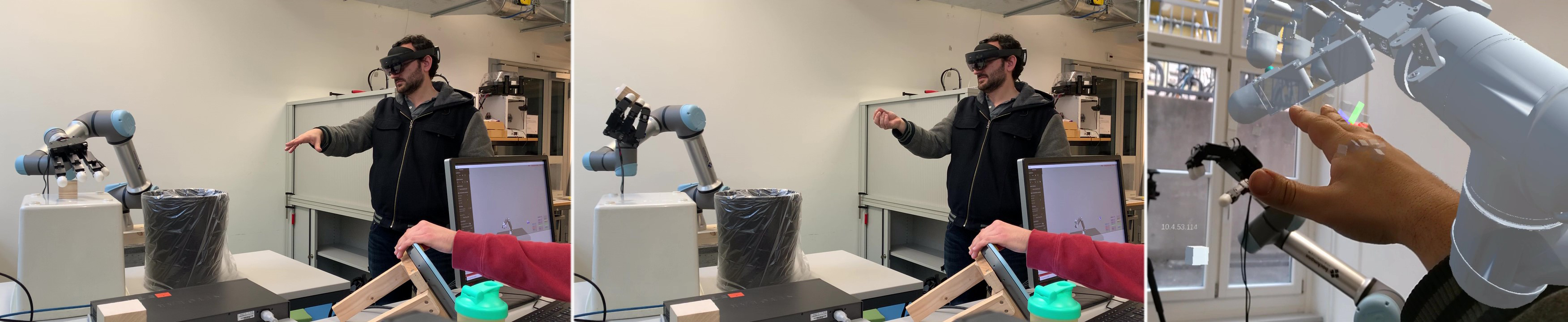}
    \caption{Immersive teleoperation by embodiment of a system comprising of an Allegro hand mounted on a UR5. The system tracks the motions of the operator shown in the image using a Hololens. This data is continuously retargetted and submitted to the robot. (Left) the operator moves into a pose that will cause the arm to grab the block on the table. (Middle) The operator changes pose to lift the cube. (Right) Inside view of the Hololens, showing the robot hologram superimposed over the user's hand, with the real robot in the background.}
    \label{fig:teleop_ur}
\end{figure}

We conducted experiments to evaluate these retargeting methods, and found that the retargeting method of choice, and the ideal parameters for it, change drastically
between users, to the point where the choice of one user is absolutely unusable by another.
Retargeting does not result in what they perceive as the correct motion they aimed to perform, and thus it does not lead to the desired degree of body ownership.
One of the main reason is that simplistic retargeting approaches do not faithfully convey the operator's intent.
The setup used to test this is shown in Fig. \ref{fig:teleop_ur} where we use a system comprising of an Allegro hand mounted on a UR5.
The operator could see a holographic digital twin of the system, overlayed over their hand, which provides some visual feedback, as well as the real robot in the background.
The users were asked to manipulate the hand and perform a pick-and-place task, with different objectives and parameters applied.
Following the experiment, users were asked to describe their experience, including sources of frustration in operating the system.
In the ensuing discussions, it became evident that different users have widely differing preferences for the retargeting model.
The hypothesis is that personalizing the model to the user will have them perform tasks faster and more accurately and will reduce mental strain (evaluated using, e.g., NASA-TLX).

The HoloLens provides a new and exciting methodology for experimenting and optimizing the user experience of retargeting.
This is thanks to the ability to overlay a hologram of the robot directly over the operator's own body, both sharing the same frame of reference, in contrast to the operator watching the real robot react to their motions from an entirely different perspective, which we observed to be quite confusing for many users.
The methodology we propose for user-specific retargeting is as follows:
1) Users are shows different robot poses as holograms;
2) they then attempt to pose themselves to match the robot poses;
3) then, given a retargeting multi-objective, we can solve an optimization problem that finds the weights that reproduces the users' poses.
This involves solving a bi-level optimization problem that can be done, for example, using sensitivity analysis~\cite{zimmermann2020}.
A preliminary study showed that there is indeed variability to the set of weights optimized for each user.
It remains to be seen what the exact distribution of weights is, across different users, and whether single \emph{average} set will be effective.

\subsection{Task Autocorrection}
Despite the many advances, demonstrations show that state-of-the art systems, while impressive, are still far from exhibiting human dexterity, and it is clear that a totally immersive experience will not be attainable in the foreseeable future.
One of the issues that will certainly linger is the difference between human and robot morphology.
Even with personalized retargeting, this difference will inhibit the sense of body ownership for the time being.
However, robots can already autonomously perform many simple tasks, such as pick-and-place, that human operators still struggle with.
The question is therefore, how to bridge this gap?
One approach is known as \emph{Assistive Teleoperation}.
As described in~\cite{Dragan-2012-7532}, it is the process of \emph{arbitrating}, i.e., \emph{blending}, the operator's motion with a learned optimal policy for a specific task.
Thus, if the task is known in advance, the system can retarget the motions by making subtle tweaks to the user's inputs, so that the motion successfully performs the task and the modifications are not too intrusive for the user.
We build on this concept and propose a new concept which we term \emph{Task Autocorrection}, which is similar but more geared towards immersiveness.
Task autocorrection refers to the process of making small modifications to the movements of the avatar, such that they match the intention of the operator.

The autocorrection framework consists of several components: In addition to the teleoperation setup, the framework consists of a retargeter that maps input motions to the avatar, an \emph{intent predictor} that outputs the user's intent, and the \emph{task corrector} that adapts the retargeter based on the predicted intent.
The retargeter and predictor provide continuous output as the user controls the robot.
The task corrector computes the optimal robot motion for the predicted task and blends the user's motion with the optimal one, based on the confidence level of the prediction.

As a proof of concept, we devised a simple deep learning-based autocorrection prototype for the holographic teleoperation setup described above.
In this system, the operator uses a HoloLens 2 to manipulate a holographic robotic arm in a scene consisting of the robot, two desks, and several balls (Fig. \ref{fig:autocorrect_scene}). 
The tasks that were considered were picking up and placing the balls and sliding a grasped ball between two walls.
The input data for training the intent predictor was the hand palm trajectory, the trajectories of the 26 hand joints, the gaze (origin and direction) trajectory, and the stacked scene data.
These were labeled by the action itself (e.g., grab or release) and the object it was applied to (e.g., blue ball).
The input is transformed into four equal-sized embedded vectors by forward propagation along two stacked Long Short Term Memory (LSTM) layers followed by one ReLU layer in each branch.
The embedding vectors are fused together by learnable weighted averaging and then fed into a final ReLU network before computing the soft-max output.
The output is a vector, which represents the probability for each action.
For the prototype, the actions were \textit{Pick}, \textit{Place}, \textit{Slide-Between-Walls} and \textit{None}.
In case a pick or place action were predicted, the target output which object the action was applied to.
The training dataset was generated by recording users performing these tasks, e.g., pick up a ball and place it on a desk randomly or pick up a ball and slide it between two walls. 
Around 5 hours of training data were gathered.
The trained model achieved 89.48\,\% accuracy for action prediction and 85.02\,\% accuracy for target prediction on the test set.

Next, for each action, an optimal trajectory was defined. 
For example, to place a ball on a desk, a smooth motion that does not penetrate the desk must be computed.
Sliding a ball through the between the two walls must be done in a perfect linear motion.
The optimal trajectory then arbitrates the operator's motion based on the confidence level and an aggressiveness factor that is determined experimentally.

\subsection{Autocorrection User Study}
\begin{figure}
    \centering
    \begin{subfigure}[t][6cm][t]{0.475\textwidth}
        \centering
        \includegraphics[height=6cm]{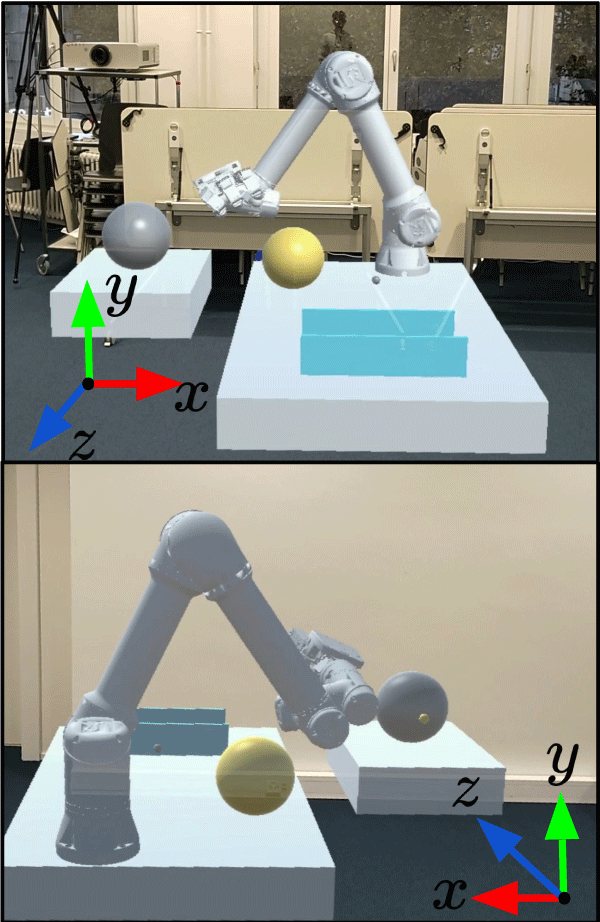}
        \caption{The scene}
        \label{fig:autocorrect_scene}
    \end{subfigure}
    \hfill
    \begin{subfigure}[t][6cm][t]{0.475\textwidth}
        \centering
        \includegraphics[height=6cm]{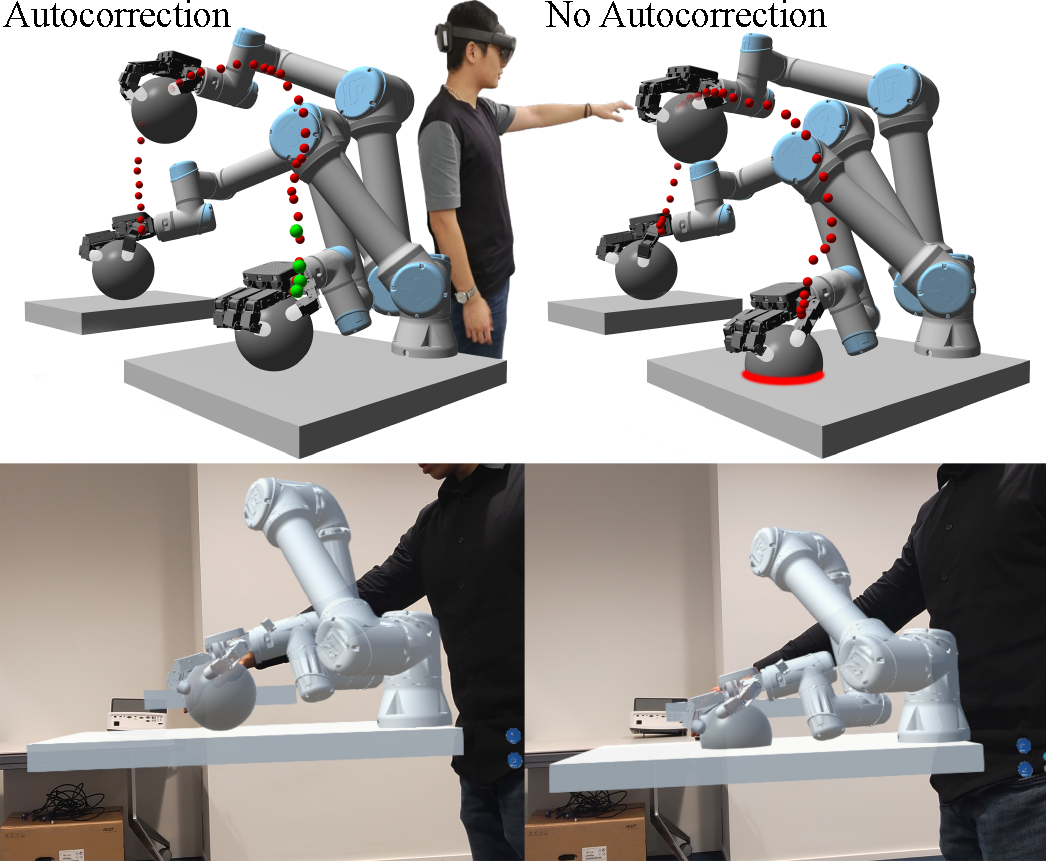}
        \caption{Autocorrection on/off}
        \label{fig:autocorrection_demo}
    \end{subfigure}
    \vspace{0.4cm}
   \caption{Autocorrection makes it easier for user to perform tasks accurately. As seen in the right image, without autocorrection, user might place the ball through the table. With auocorrection, the system identifies the task and assists with the optimal motion.}
    \label{fig:autocorrection}
\end{figure}

To evaluate the performance of the prototype, we held a small user study (n=7), where participants were invited to interact with our system, with or without the assistance of autocorrection.
Similar user studies have compared the usability of several non-MR teleoperation methods~\cite{kent2017comparison}, and provided useful insights into the effect of the user interface on the effectiveness and ease of use in manipulation tasks.
For other tasks such as robotic inspection, augmented reality interfaces have demonstrated significant benefits for users in the speed and performance of robot teleoperation~\cite{hedayati2018improving}.
The participants were required to grab the grey ball first and place it on a spot marked by an indicator (shown as a small sphere in the hologram) and then grab the yellow ball and place it next to another indicator.
The participants were told to try to place the ball \emph{properly}, that is, it should not be placed in the air or through the surface of the desk. 
As seen in Fig. \ref{fig:autocorrection_demo}, without autocorrection, due to the operator's limited capacity to control the robot, the ball is placed through the desk, but with autocorrection enabled, it is placed perfectly.

In a separate experiment, participants were asked to pick the gray ball first and drag it along the slot from side to side, twice.
In this case, the participants were told to keep the ball as low as possible without colliding with the desk or the walls.
Before starting the experiment itself, a warm-up trial was conducted to let the participants become familiar with basic operations.
For each trial, the participants were asked to perform either a \textit{Pick-And-Place} task or a \textit{Slide} task once with or without enabling autocorrection, and without them knowing whether it enabled or not.

The recorded experiments were evaluated using two quantitative metrics during simulation: \textit{Violation Distance} and \textit{Operation time}.
In addition, the participants were asked to rate the degree of \emph{naturalness}, from 1 being ``very unnatural'' to 5 being ``very natural'' during control for each trial.
A Task Load Index (TLX) was also rated by participants for this question referring to the \textit{Effort} rating of the NASA Task Load Index (NASA-TLX), where we rescaled it from 1, meaning ``Very easy”, to 5 meaning ``Very hard”.

The hypotheses were that the autocorrected motions will reduce the violation distance, shorten the operation time, and lower the TLX, while lowering the naturalness score as well.
Analysis shows that the improvement is statistically significant, with the downside of the reduced naturalness score.
This drawback could be addressed by using a more refined autocorrection system.

\subsection{Outlook}
The solutions presented in this section address some of the challenges with remote teleoperation of robot arms.
Motion retargeting and task autocorrection make the human-robot interaction \emph{feel} more natural to the user, and behave more naturally by assisting the user in executing accurate motions.

Looking forward, there are several limitations to the current framework.
The mixed reality device acts as an egocentric sensor platform, capturing the human's actions, but these works do not yet take advantage of either the robot's or the human's spatial context.
Extending the proposed system to tasks that require dynamics will require physical simulation.
A realistic simulation will further enable gathering data that includes different grasps and more dexterous manipulations tasks.
Considering its scalability to many tasks, another limitation of the proposed approach is the necessity to obtain data for each task and define each optimal trajectory.
To this end, a database of task and optimal trajectory must be established, and autocorrection could even be used to bootstrap it.
In addition to improving the generalization of these methods to more, and more arbitrary, tasks, future work to leverage spatial computing will provide these immersive and embodied teleoperation experiences with more capabilities to accomplish tasks in real-world environments.
Separate from the addition of new features to the system, an expanded user study with more participants and deeper experiments would provide valuable information to guide future work in preparing autocorrection to be applied to real-world tasks.

%% file: conclusion.tex
\section{Conclusion}
This paper has presented several prototype systems that utilize robots and mixed reality devices to provide novel solutions to compelling real-world applications through human-robot interaction.
The two key technologies that enable these solutions are the spatial computing and egocentric sensing capabilities of mixed reality devices.
All three systems make use of one or both of these to provide new spatial capabilities, as well as intuitive and natural interaction.
However, these systems are only preliminary explorations toward these real-world applications, and our initial results have uncovered more new questions than they have answered.

The primary barrier to practical deployment of the proposed mission planning framework from Sec.~\ref{sec:mission_planning} is the lack of support for large-scale, continuous maps in Azure Spatial Anchors (or any other current spatial anchor cloud provider).
Ideally, we would like to have our robots and devices localized continuously as they move through the environment, not just when they are near to a particular anchor.
This capability will require large-scale, continuous maps in the cloud, and a service that provides localization to a venue (rather than an individual anchor) that is robust to the network dropouts that may occur in large indoor spaces. 
Both the robots and MR devices should also be able to leverage a priori spatial information from Building Information Models (BIM) and CAD data to visualize and use both as-planned information from these designs and as-built information from physically being on site.
These features are not yet present in current commercial localization services, but state of the art research has demonstrated that these capabilities are ready to be applied to these real-world scenarios.

On the other hand, there are still many open avenues for research in human-robot interaction through MR.
In particular, although we have demonstrated intuitive control by giving a robot commands through natural hand gestures, the robot's understanding of this interaction is no different than if these navigation goals were provided by a keyboard and mouse, and its representation of its environment is still one of obstacles and free space.
While this lack of semantic understanding is not a barrier for simple tasks like navigation to a goal, further work will be required to endow the robot with the understanding of human presence and the semantic context necessary to perform more complex tasks that can be communicated in a natural way by humans.
We have released several pieces of open source software (see Sec.~\ref{sec:mission_planning}) to help facilitate research in this domain that interfaces HoloLens 2 devices and robots.

Within immersive teleoperation, some of the major open questions will need to be resolved through better human understanding.  
The user studies around motion retargeting and task autocorrection have so far been limited in scope.
Consequently, there are questions about whether it is possible and beneficial to provide personalized models for each user to provide an improved experience from a single method.
Deeper and broader user studies should be performed to better understand the requirements for general usability, and to provide insights on future research questions to pursue.

Bringing both spatial and human understanding together in a human-robot system offers promising new modes of interaction.  
We have shown several mixed reality-based systems that leverage these technologies to provide solutions to real-world robot use cases.  
While these capabilities are deployable today, and applicable to many commercial and research scenarios beyond the scope of this paper, these works represent only the tip of the iceberg, with many possibilities for future research in spatial computing and human-robot interaction.

%% file: acknowledgment.tex
\section*{Acknowledgment}
The authors would like to thank Oswaldo Ferro, Jan St\"{u}hmer, Lukas Gruber, and Simon Huber for their contributions to these projects.